%% file: acl_latex.tex
\documentclass[11pt]{article}

\usepackage[preprint]{acl}

\usepackage{chngcntr} 
\usepackage{graphicx}
\usepackage{times}
\usepackage{latexsym}
\usepackage{hyperref}
\usepackage{amsmath}
\usepackage{booktabs}
\usepackage{stfloats}
\usepackage{multirow}
\usepackage{tabularx}
\usepackage{graphicx}
\usepackage{hyperref}
\usepackage{cleveref}
\usepackage{float}
\usepackage[T1]{fontenc}

\usepackage{algorithm}
\usepackage{algorithmic}
\usepackage{amsmath}
\usepackage{booktabs}
\usepackage{multirow}
\usepackage{array}

\usepackage[utf8]{inputenc}

\usepackage{microtype}

\usepackage{inconsolata}

\usepackage{graphicx}
\usepackage[most]{tcolorbox}
\tcbuselibrary{listings}

\newtcblisting[auto counter, number within=section]{promptbox}[2][]{
 enhanced,
 breakable,
 title={\textbf{} #2},
 colback=gray!5,
 colframe=gray!60,
 colbacktitle=gray!30,
 coltitle=black,
 fonttitle=\bfseries\small,
 attach boxed title to top left={xshift=3mm, yshift=-3mm},
 boxed title style={size=small, colframe=gray!60},
 sharp corners,
 top=4mm,
 bottom=2mm,
 left=2mm, right=2mm,
 listing only,
 listing options={
   basicstyle=\small\ttfamily,
   breaklines=true,
   columns=fullflexible,
   keepspaces=true,
   aboveskip=0pt, belowskip=0pt
 },
 label={#1}
}

\newcommand{\ours}{MemCompile}
\newcommand{\best}[1]{\textbf{#1}}
\newcommand{\second}[1]{\underline{#1}}

\title{From Memory Hits to Answerable Evidence: Query-Conditioned Memory Compilation for Long-Term Agent Memory}
\author{
  \textbf{Shangze Li\textsuperscript{1}},
  \textbf{Yanyu Yao\textsuperscript{1}},
  \textbf{Derong Xu\textsuperscript{1}},
  \textbf{Zhi Zheng\textsuperscript{1}},
  \textbf{Tong Xu\textsuperscript{1}}
  \\
\textsuperscript{1}State Key Laboratory of Cognitive Intelligence, \\University of Science and Technology of China, Hefei, China
\\
\\
\small{
\href{mailto:lishangze@mail.ustc.edu.cn, yyyao@mail.ustc.edu.cn, derongxu@mail.ustc.edu.cn}{\{lishangze ,yyyao, derongxu\}@mail.ustc.edu.cn}, \href{mailto:zhengzhi97@ustc.edu.cn,tongxu@ustc.edu.cn}{\{zhengzhi97, tongxu\}@ustc.edu.cn}
  }
}

\begin{document}

\maketitle

\begin{abstract}

Long-term agent memory requires recovering answer-relevant evidence dispersed across turns and sessions. Existing systems increasingly use structured memories and adaptive retrieval, yet often pass retrieved or expanded units directly to the reader, coupling memory addressing with evidence presentation and leaving the final context redundant or incomplete. To address this gap, we propose \ours{}, a query-conditioned evidence compilation framework for long-term conversational memory. Given a query, the Address Plane locates relevant support through compact source-linked descriptors and capacity-aware Cells, while a normalization boundary resolves heterogeneous retrieval hits to stable SourcePacket identities. Its Evidence Plane then allocates a bounded evidence budget across complementary packets, admits limited additional support, and renders the retained claims associated with each selected packet into an answer-ready context. \ours{} achieves 80.52\% Judge Accuracy on LoCoMo and extends the empirical quality–cost Pareto frontier to a higher-quality regime. Its strong performance on LongMemEval-S further suggests that the benefits of query-conditioned compilation generalize across long-term conversational-memory benchmarks.
The implementation code is publicly available at \url{https://github.com/MINE-USTC/MemCompile}

\end{abstract}

\begin{figure}[!t]
\centering
\includegraphics[width=\columnwidth]{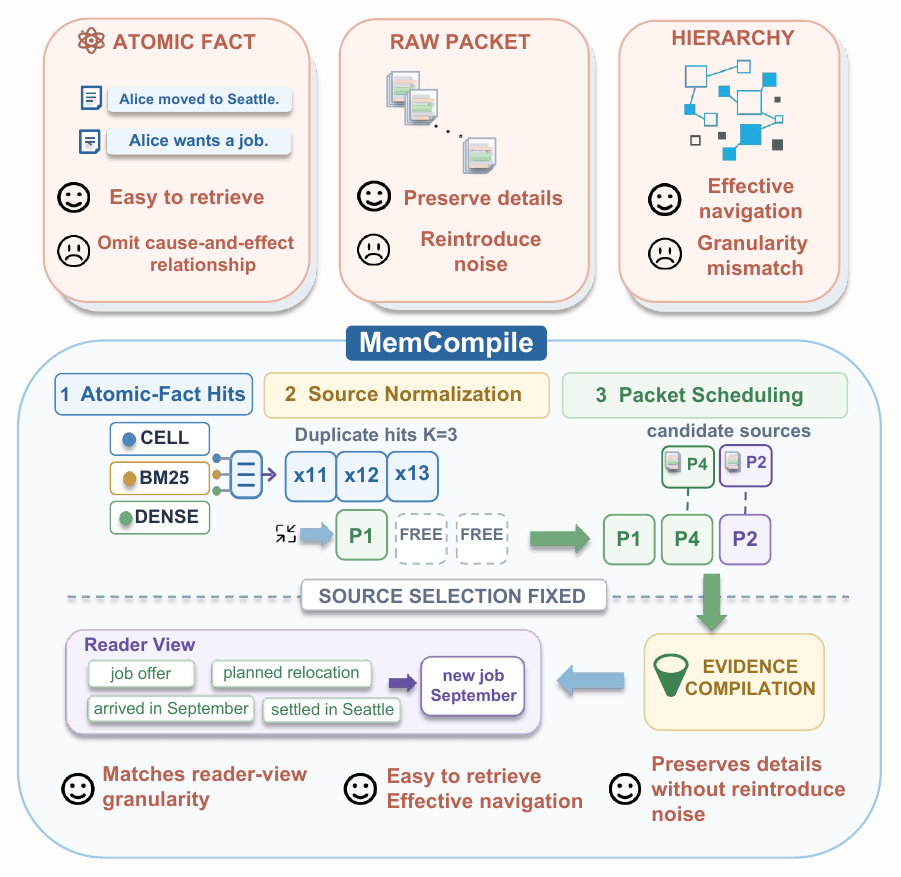}
\caption{\textbf{Motivation and overview of MemCompile}.
    Existing memory systems face different trade-offs in retrieval units
    and evidence representation. MemCompile addresses these challenges
    through source-normalized retrieval, complementary evidence selection,
    and packet-scoped reader-view compilation.}
\label{fig:motivation}
\end{figure}

\section{Introduction}

Long-term memory is essential for interactive agents to sustain coherent, personalized, and reliable behavior across sessions. By carrying forward user preferences, past events, commitments, and evolving states, it allows an agent to respond in light of what has happened before \cite{locomo, wu2024longmemeval}. However, the information needed for a new query may be distributed across distant turns and sessions or revised by later interactions. Persistent interaction therefore depends not only on retaining past information, but also on making the relevant parts of an accumulated history accessible when needed.


Foundational long-term memory systems addressed this need by extracting and retrieving salient records, including persistent memories and atomic conversational facts \cite{memorybank, mem0}. Subsequent work improved access by compressing dialogue into facts, notes, summaries, profiles, and events, and by organizing these representations in graphs or hierarchies \cite{Amem, zep, simplemem, Atommem}. As retrieval representations became increasingly compact and structured, recent work began to distinguish the units used to locate memory from the evidence exposed to the reader. AnchorMem reconstructs immutable contexts from retrieved fact anchors, xMemory adaptively discloses segments or raw messages from a hierarchy, and MemIR projects heterogeneous retrieval cues into provenance-scoped claim bundles \cite{anchormem, xmemory, MemIR}. Together, these developments move long-term memory systems from improving addressability toward separating memory location from evidence presentation.



However, this separation remains incomplete: the retrieved representation, or the expansion attached to it, still largely determines the context presented to the reader. Existing systems therefore do not fully separate two decisions: which underlying conversation sources should receive the limited evidence budget, and how evidence from those sources should be represented for answer generation. This coupling is problematic because the two decisions favor different properties. Address representations should be compact, discriminative, and broad enough to support reliable recall, whereas reader evidence should preserve relevant qualifications, connect complementary facts, and avoid unnecessary context. When a single unit serves both purposes, the system must accept a fixed trade-off. As illustrated in Figure~\ref{fig:motivation}, an atomic claim may be easy to retrieve but may omit causal relationships or sibling facts from the same source.  A raw turn preserves such details but may reintroduce noise, such as greetings and irrelevant dialogue, whereas a hierarchy node can provide an effective navigation unit without matching the granularity required for answer generation. Moreover, without an explicit source-level boundary, multiple claims or retrieval channels may allocate context budget repeatedly to the same underlying source. This coupling not only creates avoidable quality loss after successful retrieval, but also makes it difficult to distinguish failures of source recovery from failures of evidence presentation.

In this paper, we propose \ours{}, a source-normalized evidence compiler for long-term conversational memory. Its central principle is to locate relevant information through compact representations and allocate the evidence budget at the level of original dialogue turns while generating from a separately compiled reader view. \ours{} realizes this principle through an Address–Evidence two-plane architecture. The Address Plane organizes source-linked atomic descriptors into capacity-bounded Cells and combines diminishing-benefit Cell routing with lexical retrieval to construct primary source candidates while a broader dense packet pool supplies proposals for bounded semantic refinement. An explicit normalization boundary resolves these heterogeneous results to stable SourcePacket identities, ensuring that one underlying source occupies one evidence unit regardless of how many descriptors or channels locate it. The Evidence Plane then selects complementary packet identities, preserves the initial anchors while admitting bounded additional support, completes evidence within each selected packet, and compiles the result into a reader-facing view. By separating packet selection from textual rendering, \ours{} can optimize and evaluate source recovery and evidence presentation as distinct decisions.


Our main contributions are summarized as follows:
\begin{itemize}
\item We formulate long-term memory serving as query-conditioned evidence compilation, distinguishing two decisions that are commonly coupled: which underlying sources receive the evidence budget and how evidence from those sources is represented to the reader. 
\item We introduce \ours{}, an Address–Evidence two-plane architecture that addresses memory through compact derived views, allocates evidence budget at the level of immutable SourcePackets, and generates from a separately compiled reader view. 
\item Comprehensive evaluations on long-term memory
benchmarks demonstrate that \ours{} consistently outperforms state-of-the-art baselines while  extending the empirical quality–cost Pareto frontier. On the LoCoMo benchmark,  packet-scoped compilation improves F1 by 15.26 points and Judge Accuracy by 23.51 points over representative-claim rendering.
\end{itemize}

\section{Related Work}
\paragraph{Structured long-term memory for LLM agents}
Early long-term memory systems store and retrieve salient interaction records to support persistent personalization and multi-session reasoning. MemoryBank combines retrieval with time-dependent forgetting and memory  reinforcement, while Mem0 extracts and consolidates conversational facts into vector- and graph-based stores \cite{memorybank, mem0}. Subsequent systems construct richer memory organizations. A-MEM creates dynamically linked notes whose attributes and relations evolve as new experiences arrive while Zep represents evolving conversational knowledge in a temporal graph \cite{Amem, zep}. Other approaches improve information density or associative access through
semantic compression, atomic facts, profiles, events, and higher-level memory scenes \cite{simplemem, Atommem}.

\paragraph{Reconstructive and adaptive memory access}
Recent systems recognize that a representation useful for locating memory need not be self-contained answer evidence. AnchorMem retrieves atomic facts as anchors and reconstructs their associated raw contexts and events \cite{anchormem}. xMemory organizes intact memory units into a revisable hierarchy and adaptively discloses episodes or raw messages when additional
detail reduces reader uncertainty \cite{xmemory}. MemIR takes a typed approach, separating raw evidence, retrieval cues, and truth-bearing claims, and projecting heterogeneous hits into provenance-scoped claim bundles exposed through a normalized fact
interface \cite{MemIR}. \ours{} builds on the shared insight that retrieval representations and answer evidence serve different purposes, but places its architectural boundary at a different level. \ours{} instead resolves derived
address results to immutable SourcePacket identities before evidence selection. This source-normalized boundary makes it possible to vary and evaluate reader representation while holding the selected source set fixed.

\section{Methodology}
\label{sec:method}

\subsection{Problem Formulation}
\label{sec:problem-formulation}

Let a conversational history be
$\mathcal{H}=(\mathcal{P},\prec)$, where
$\mathcal{P}=\{p_1,\ldots,p_N\}$ contains the original dialogue turns
and $\prec$ denotes their temporal order. Each turn $p_i$ includes its
speaker, session, interaction time, and content. Given a query $q$, the
task is to construct an ordered and bounded reader context $X_q$ grounded
in $\mathcal{H}$. The context should contain sufficient and complementary
support for answering $q$, preserve traceability to the original dialogue
turns, and limit irrelevant or redundant content.

\subsection{Address Plane}
\label{sec:address-plane}
\ours{} instantiates each original turn $p_i\in\mathcal P$ as an
immutable SourcePacket that stores its speaker, session, interaction
time, raw content, and stable identifier. The Address Plane builds compact source-linked descriptors and organizes them for complementary retrieval. Its objective is to preserve source coverage under fixed limits on opened Cells and sparse and dense candidate pools.

\begin{figure*}[t]
\centering
\includegraphics[width=1.03\textwidth]{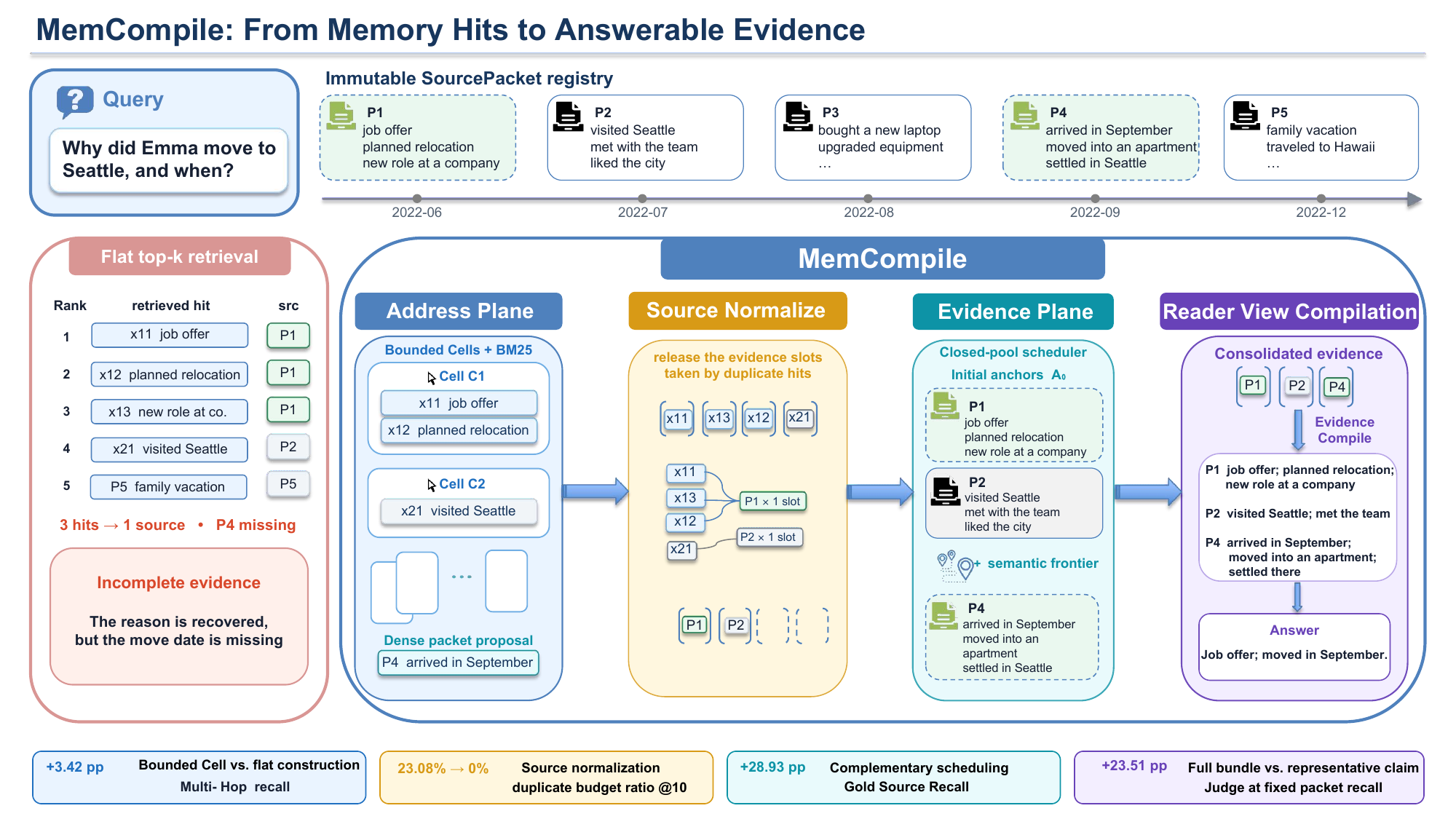}
\caption{\textbf{The overall architecture of \ours{}}}
\label{fig:Overview}
\end{figure*}

\subsubsection{Source-Linked Atomic Descriptors}
\label{sec:atomic-descriptors}
From immutable SourcePackets, \ours{} constructs the descriptor
set $\mathcal X=\{x_1,\ldots,x_M\}$. Each descriptor retains a foreign key to its owning SourcePacket,
\begin{equation}
  \pi:\mathcal X\rightarrow\mathcal P.
\end{equation}
For each SourcePacket, a large language model extracts standalone factual claims while
using the immediately preceding turn to resolve local references.  A retained
descriptor is
\begin{equation}
 \begin{aligned}
 x_i=\bigl(f_i,d_i,s_i,j_i,t_i,\mathsf{Per}_i,\mathsf{Key}_i,
              \tau_i,\bar{\tau}_i,n_i,\pi_i\bigr),
 \end{aligned}
 \label{eq:atomic-descriptor}
\end{equation}
where $f_i$ is the claim text, $d_i$ is the dialogue-turn identifier, $s_i$
is the speaker, $j_i$ is the session index, and $t_i$ is the recorded session
time.  $\mathsf{Per}_i$ and $\mathsf{Key}_i$ are the extracted people and
keyword sets; $\tau_i$ and $\bar{\tau}_i$ are the described-event time and
interaction time; $n_i$ is the claim token count; and
$\pi_i\in\mathcal{P}$ is its SourcePacket foreign key.  Claims provide sparse
and semantic retrieval keys while preserving a path to their originating
dialogue evidence.

\subsubsection{Bounded Cell Placement}
\label{sec:cell-placement}
Cells are capacity-constrained routing structures over atomic descriptors.
Let $\mathcal{C}$ be the set of Cells. Descriptor $x_i$ is feasible for cell $c$ only if
\begin{equation}
  \operatorname{Token}(c)+n_i\le B_C,
  \qquad |c|<M_C,
  \label{eq:cell-capacity}
\end{equation}
where $B_C$ is the token capacity and $M_C$ is the item capacity. A feasible
Cell is scored by
\begin{equation}
 \begin{aligned}
  S_{\mathrm{place}}(i,c)={}&L(i,c)+\beta C(i,c)\\
  &{}-\gamma X(i,c)-\eta E(i,c),
 \end{aligned}
  \label{eq:cell-placement-score}
\end{equation}
where $L$, $C$, $X$, and $E$ respectively measure metadata locality, weighted
keyword--person--time coherence, polarity inconsistency, and capacity/entropy
pressure. Creating a new Cell is an additional candidate with expansion
pressure $0.65+n_i/B_C$. The highest-scoring candidate receives $x_i$.
Placement is serialized in dialogue and claim order, producing a fixed Cell
layout before query serving. Appendix~\ref{app:cell}
specifies every score component, hyperparameter, and stopping condition.

\subsubsection{Diminishing-Benefit Cell Routing}
\label{sec:cell-opening}

For a query $q$, we calculate the query relevance $u(c,q)$ for each cell based on the normalized query term set of $q$ and the person set.

Given the set of memory units $\mathcal O_t$ opened before round $t$, the marginal benefit of candidate memory cell $c$ is denoted as $g(c\mid\mathcal O_t,q)$:

\begin{equation}
\label{eq:cell-marginal-gain}
g(c\mid\mathcal O_t,q)
=
u(c,q)
\left[
1-\lambda_R
J\!\left(M_c,M_{\mathcal O_t}\right)
\right]
-
\kappa_{\mathrm{open}}
\end{equation}

Here, $u(c,q)$ combines keyword, person, time, and density signals and $J(A,B)$ denotes Jaccard similarity between two sets.
$M_c$ contains the Cell's people and keywords, and $M_{\mathcal O_t}$ is their
cumulative set over opened Cells. The implementation sets $\lambda_R=0.45$
and defines $\kappa_{\mathrm{open}}$ as an opening cost that increases with the number of
opened Cells plus a token-proportional reading cost. Cells are opened greedily
until the marginal gain or relative-utility criterion fails, the cumulative
utility reaches its threshold, or the six-Cell cap is reached. Exact
weights and stopping conditions appear in Appendix~\ref{app:cell-routing}.

\subsubsection{Multi-Channel Candidate Union}
\label{sec:candidate-union}

BM25 ranks all atomic descriptors and supplies a lexical channel alongside
Cell routing; its top eight descriptors also act as Cell-routing seeds. Let
$\mathcal{C}_q$ denote the opened Cell set, the complete address candidate pool is
\begin{equation}
 \begin{aligned}
  \mathcal{K}_q={}&
  \bigcup_{c\in\mathcal{C}_q}\operatorname{claims}(c){}\cup \operatorname{BM25}(q),
 \end{aligned}
  \label{eq:candidate-union}
\end{equation}
where $\operatorname{BM25}(q)$ denotes the top 96 lexical candidates.
For $x_i\in\mathcal{K}_q$, let $\mathsf{Term}_{i}$ be the normalized claim
terms and let $b_i$ be its normalized BM25 score, or zero when it is not
returned by the lexical channel. Its candidate score is
\begin{equation}
 \begin{aligned}
  r_i={}&\alpha_fJ(\mathsf{Term}_q,\mathsf{Term}_{i})
  +\alpha_kJ(\mathsf{Term}_q,\mathsf{Key}_i)\\
  &+\alpha_pJ(\mathsf{Per}_q,\mathsf{Per}_i)
  +\alpha_t\mathbf{1}_t(i,q)+\alpha_b b_i,
 \end{aligned}
  \label{eq:claim-score}
\end{equation}
where $\mathbf{1}_t(i,q)$ indicates that a temporal query matches a descriptor
with normalized time. The coefficients are specified in
Appendix~\ref{app:cell-routing}. The descriptors in the union are then
reordered according to the candidate score.

In parallel with structured Cell routing, the supplemental channel performs global semantic retrieval over the complete packet collection. Let $K_D$ denote the semantic proposal depth. It encodes the enriched query $q^+$ and each packet document $z(p)$, and selects the $K_D$ packets with the highest
cosine similarity; we use $K_D=40$ in the reported configuration:
\[
\mathcal D_q
=
\operatorname{TopK}^{K_D}_{p\in\mathcal P}
\operatorname{cos}\!\left(e(q^+),e(z(p))\right).
\]

\subsection{Plane Transition via Source-Normalization Boundary}
\label{sec:plane-transition}

The candidate pool $\mathcal{K}_q$ contains address hits rather than distinct
pieces of evidence.  Several claims can paraphrase the same dialogue turn,
and the same claim can be reached through both BM25 and one or more Cells.  We
therefore resolve every claim through
\begin{equation}
  \rho(x_i)=\pi_i\in\mathcal{P}
  \label{eq:claim-resolution}
\end{equation}
and collapse all candidates with the same resolved packet.  The normalized
packet pool is
\begin{equation}
 \begin{aligned}
  \mathcal{U}_q=\bigl\{p\in\mathcal{P}:{}&
    \exists x_i\in\mathcal{K}_q,\rho(x_i)=p\bigr\},
 \end{aligned}
  \label{eq:packet-pool}
\end{equation}
with packet score
\begin{equation}
  s(p,q)=
  \max_{\substack{x_i\in\mathcal{K}_q\\\rho(x_i)=p}} r_i.
  \label{eq:packet-score}
\end{equation}
This many-to-one transition
separates the multiplicity of retrieval keys from the amount of evidence:
multiple successful addresses to one turn create one evidence candidate, not
multiple budget items.

\subsection{Evidence Plane}
\label{sec:evidence-plane}

The Evidence Plane converts the normalized packet candidates into a bounded
and coherent context. We organize it around the two decisions: how to compile a selected packet,
and when to supplement the selected packet set.
\subsubsection{Complementary Packet Scheduling }
A closed-pool listwise scheduler receives $q$ and $\mathcal{U}_q$ and selects
$K_0=10$ complementary SourcePackets,
\begin{equation}
  \mathcal{A}_0(q)
  =\operatorname{Schedule}_{K_0}(q,\mathcal{U}_q),
  \label{eq:baseline-anchors}
\end{equation}
where $\mathcal{A}_0(q)$ is the baseline evidence anchor set. Each candidate is
rendered with an opaque label, dialogue identifier, speaker, recorded date,
and source content. The scheduler is instructed to rank direct answer-bearing
packets first while retaining complementary packets for relational, temporal,
comparative, and list questions. It returns exactly $K_0$ distinct labels in
descending usefulness; labels are validated against the closed pool and
resolved back to SourcePacket identities. The scheduler can therefore reorder
and select only packets in $\mathcal{U}_q$. Appendix~\ref{app:scheduler}
provides the complete request construction, output validation, and prompt.
After this transition,
Cells, claim multiplicity, and retrieval-channel identity no longer define the
evidence state; the Evidence Plane operates over SourcePackets and their
reader projections.

\subsubsection{Anchor-Preserving Bounded Frontier}
\label{sec:evidence-frontier}

Source-scoped claim-bundle rendering improves the usability of selected evidence, while
the frontier addresses support that is absent from the baseline anchor set.  The closed frontier pool is
\begin{equation}
  \mathcal{W}_q
  =\mathcal{A}_0(q)\cup\mathcal D_q.
  \label{eq:frontier-pool}
\end{equation}
A listwise reranker returns an ordering $\sigma_q$ over $\mathcal{W}_q$.  The
system examines its first $K_R=5$ positions and admits at most $K_\Delta=2$
previously unselected packets:
\begin{align}
  \Delta(q)
  &=\operatorname{First}_{K_\Delta = 2}
    \left(\sigma_q[1{:}K_R]\setminus\mathcal{A}_0(q)\right),
  \label{eq:frontier-delta}\\
  \mathcal{E}_q
  &=\mathcal{A}_0(q)\cup\Delta(q).
  \label{eq:final-evidence-set}
\end{align}
Here $\operatorname{First}_m(S)$ returns the first at most $m$ elements of an
ordered sequence $S$ while preserving their order. The construction preserves the baseline anchors and bounds the evidence change. 
In Appendix~\ref{app:frontier-sensitivity} we report the sensitivity of reranking cutoff and append cap.

\subsubsection{Reader-View Compilation}
After obtaining the final ordered SourcePacket sequence $\mathcal{E}_q$, we
construct a source-scoped claim-bundle view for each packet $p$:
\begin{equation}
B_p=\left\langle x\in\mathcal X\mid\pi(x)=p\right\rangle_{\prec_{\mathcal X}}.
\label{eq:claim-bundle}
\end{equation}
The system retrieves every retained atomic claim originating from $p$, formats
the bundle with the packet timestamp, and concatenates the evidence blocks in
packet scheduling order. This compilation is complete with respect to the
retained descriptor set, rather than a reconstruction of the packet's raw text. Let $\mathcal E_q=\langle p_1,\ldots,p_L\rangle$ be the final scheduled
packet sequence. The final reader context is
\begin{equation}
  X_q=\operatorname{Render}(p_1,B_{p_1})\mathbin{\Vert}\cdots
  \mathbin{\Vert}\operatorname{Render}(p_L,B_{p_L}),
  \label{eq:reader-context}
\end{equation}
where $\Vert$ denotes order-preserving concatenation of the rendered evidence blocks.

\section{Experiments}
Our experiments are designed to answer four central research questions: 
\textbf{RQ1} Does MemCompile deliver meaningful performance improvements over long-term-memory baselines under the same evaluation protocol? 
\textbf{RQ2} Does bounded Cell placement improve fixed-budget source recovery, and how does it complement the Evidence Plane? \textbf{RQ3} Does budgeting by SourcePacket reduce source aliasing and distinguish source-recovery failures from evidence-presentation failures? 
\textbf{RQ4} Given the same recovered SourcePackets, how does reader-view granularity affect answer quality?
\subsection{Experimental Settings}

\paragraph{Datasets}
We evaluate \ours{} on LoCoMo \cite{locomo} and LongMemEval \cite{wu2024longmemeval} benchmarks. LoCoMo evaluates long-term conversational memory over extended, multi-session dialogues which contains ten conversations, each comprising roughly 600 dialogue turns and 16K tokens on average. LongMemEval-s evaluates the ability of chat assistants to retain and use information across sustained user-assistant interactions, which contains 500 questions, each embedded in a history of approximately 115K tokens.
\paragraph{Baselines and Metrics}
We compare against representative long-term memory systems spanning flat, atomic, graph, and hierarchical designs including MemoryBank \cite{memorybank}, A-Mem \cite{Amem}, Mem0 \cite{mem0}, MemoryOS \cite{memoryos}, LightMem \cite{fang2026lightmem}
, AnchorMem \cite{anchormem}, xMemory \cite{xmemory}
, SimpleMem \cite{simplemem}, AtomMem \cite{Atommem}.
We report token-level F1, BLEU-1 \cite{papineni-etal-2002-bleu} and Judge accuracy \cite{LLMASJUDGE}. We report token cost through memory construction and query-time retrieval/generation. For Open-Domain questions, we do not
use F1 or BLEU-1 for cross-method comparisons due to this category require combining conversational evidence
with commonsense or world knowledge and may admit multiple
lexically divergent yet semantically valid answers. 

\paragraph{Experimental Details}
Unless otherwise specified, all generative components of \ours{} use \texttt{gpt-4o-mini} with temperature $0$ and default top-$p$, and set the maximum output length to 2,048 tokens. Dense retrieval uses \texttt{text-embedding-3-small} without task-specific fine-tuning or dimensionality reduction.
The LLM-as-a-Judge metric employs \texttt{gpt-5-nano} with a rigorous evaluation prompt to assess semantic correctness.

To ensure a controlled comparison, we rerun all baselines under the same evaluation protocol while preserving their method-specific memory construction and retrieval mechanisms.  All methods are evaluated on the same dataset release and use the same model configuration including the same backbone model and the same answer/judge prompt.

\begin{table*}[t]
\centering
\scriptsize
\setlength{\tabcolsep}{1.7pt}
\resizebox{\textwidth}{!}{%
\begin{tabular}{l*{14}{c}}
\toprule
& \multicolumn{3}{c}{Single-Hop} & \multicolumn{3}{c}{Multi-Hop} & \multicolumn{3}{c}{Temporal} & Open-Domain & \multicolumn{3}{c}{Overall} & \\
\cmidrule(lr){2-4}\cmidrule(lr){5-7}\cmidrule(lr){8-10}\cmidrule(lr){11-11}\cmidrule(lr){12-14}
Method & F1$\uparrow$ & BLEU-1$\uparrow$ & J$\uparrow$ & F1$\uparrow$ & BLEU-1$\uparrow$ & J$\uparrow$ & F1$\uparrow$ & BLEU-1$\uparrow$ & J$\uparrow$ & J$\uparrow$ & F1$\uparrow$ & BLEU-1$\uparrow$ & J$\uparrow$ & Tokens (K) \\
\midrule
LoCoMo & 37.81 & 26.23 & 57.07 & 20.97 & 11.09 & 48.58 & 34.79 & 15.46 & 38.01 & 41.67 & 33.85 & 20.88 & 50.58 & 827.20 \\
MemoryBank & 17.85 & 12.13 & 46.61 & 11.25 & 8.20 & 28.01 & 14.26 & 9.60 & 34.27 & 23.96 & 15.76 & 10.80 & 39.22 & 926.98 \\
A-Mem & 33.87 & 27.64 & 54.10 & 21.17 & 15.40 & 43.26 & 34.13 & 29.98 & 34.89 & 32.29 & 31.45 & 25.77 & 46.75 & 11,687.58 \\
Mem0 & 54.95 & 44.71 & \second{82.40} & 36.02 & 25.25 & 60.99 & 30.36 & 29.69 & 33.02 & \second{50.00} & 45.79 & 37.57 & 66.17 & 55,300.30 \\
MemoryOS & 49.72 & 43.58 & 66.47 & 37.15 & 26.87 & 60.99 & 41.99 & 36.32 & 34.58 & \best{51.04} & 45.55 & 38.70 & 57.86 & 19,207.67 \\
LightMem & 49.30 & 42.45 & 71.58 & 36.59 & 27.27 & 62.41 & 47.41 & 41.29 & 57.63 & 44.79 & 46.40 & 39.23 & 65.32 & 5,021.56 \\
AnchorMem & \second{57.73} & \second{50.91} & 81.09 & 34.82 & 25.97 & 57.45 & 23.47 & 18.40 & 42.37 & 36.46 & 45.64 & 38.81 & 65.91 & 5,840.46 \\
xMemory & 50.26 & 43.81 & 81.81 & \second{40.13} & \second{31.77} & 64.89 & 53.36 & 47.33 & 60.44 & 44.79 & 48.97 & 42.24 & \second{71.95} & 17,573.40 \\
SimpleMem & 51.18 & 44.75 & 75.98 & 38.50 & 30.82 & 64.18 & 31.66 & 26.48 & 38.94 & 42.71 & 44.36 & 37.97 & 64.03 & 18,077.08 \\
AtomMem & 54.43 & 47.56 & 76.20 & 39.76 & 31.43 & \second{66.31} & \second{57.73} & \second{52.46} & \second{65.73} & 48.95 & \second{52.30} & \second{45.50} & 70.51 & 24,553.27 \\
\midrule
\ours{} & \best{59.41} & \best{51.98} & \best{88.59} & \best{41.20} & \best{32.59} & \best{75.00} & \best{59.82} & \best{54.03} & \best{73.52} & 49.48 & \best{55.94} & \best{48.65} & \best{80.52} & 26,720.82 \\
\bottomrule
\end{tabular}
}
\caption{\textbf{LoCoMo category and overall results}. Best results are bold and second-best results are underlined. Tokens exclude judging and embedding.}
\label{tab:main}
\end{table*}

\begin{table}[t]
\centering
\scriptsize
\setlength{\tabcolsep}{2.0pt}
\resizebox{\columnwidth}{!}{%
\begin{tabular}{lrrrr}
\toprule
& \multicolumn{4}{c}{Overall} \\
\cmidrule(lr){2-5}
Setting & F1$\uparrow$ & BLEU-1$\uparrow$ & Recall$\uparrow$ & Judge$\uparrow$ \\
\midrule
A0 Full system & \best{55.94} & \best{48.65} & \best{85.50} & \best{80.52} \\
A1 w/o frontier & 51.78 & 44.87 & 80.84 & 74.94 \\
A2 w/o scheduler and frontier & 42.00 & 36.01 & 51.99 & 60.91 \\
A3 w/o scheduler, with frontier & 51.91 & 44.75 & 76.55 & 75.26 \\
A4 Cell-only route & 50.64 & 43.85 & 69.85 & 71.88 \\
A5 BM25-only route & 52.31 & 45.14 & 81.54 & \second{77.53} \\
A6 w/o query intent & \second{53.61} & \second{46.47} & \second{84.84} & 77.34 \\
A7 Dense SourcePacket Top-10 & 48.80 & 42.23 & 65.69 & 70.91 \\
\bottomrule
\end{tabular}
}
\caption{\textbf{Controlled pipeline-path ablations on LoCoMo}. A1 retains the packet anchors returned by Complementary Packet Scheduling but disables subsequent semantic-frontier
expansion. A2 and A3 replace the scheduled anchors with deterministic dialogue-level MMR Top-10 anchors, the scoring function and weights are detailed in Appendix~\ref{app:cell-routing}. A7 directly selects the dense cosine Top-10
SourcePackets and presents their claim bundles to the Reader.}
\label{tab:ablations}
\end{table}

\begin{figure}[t]
\centering
\includegraphics[width=\columnwidth]{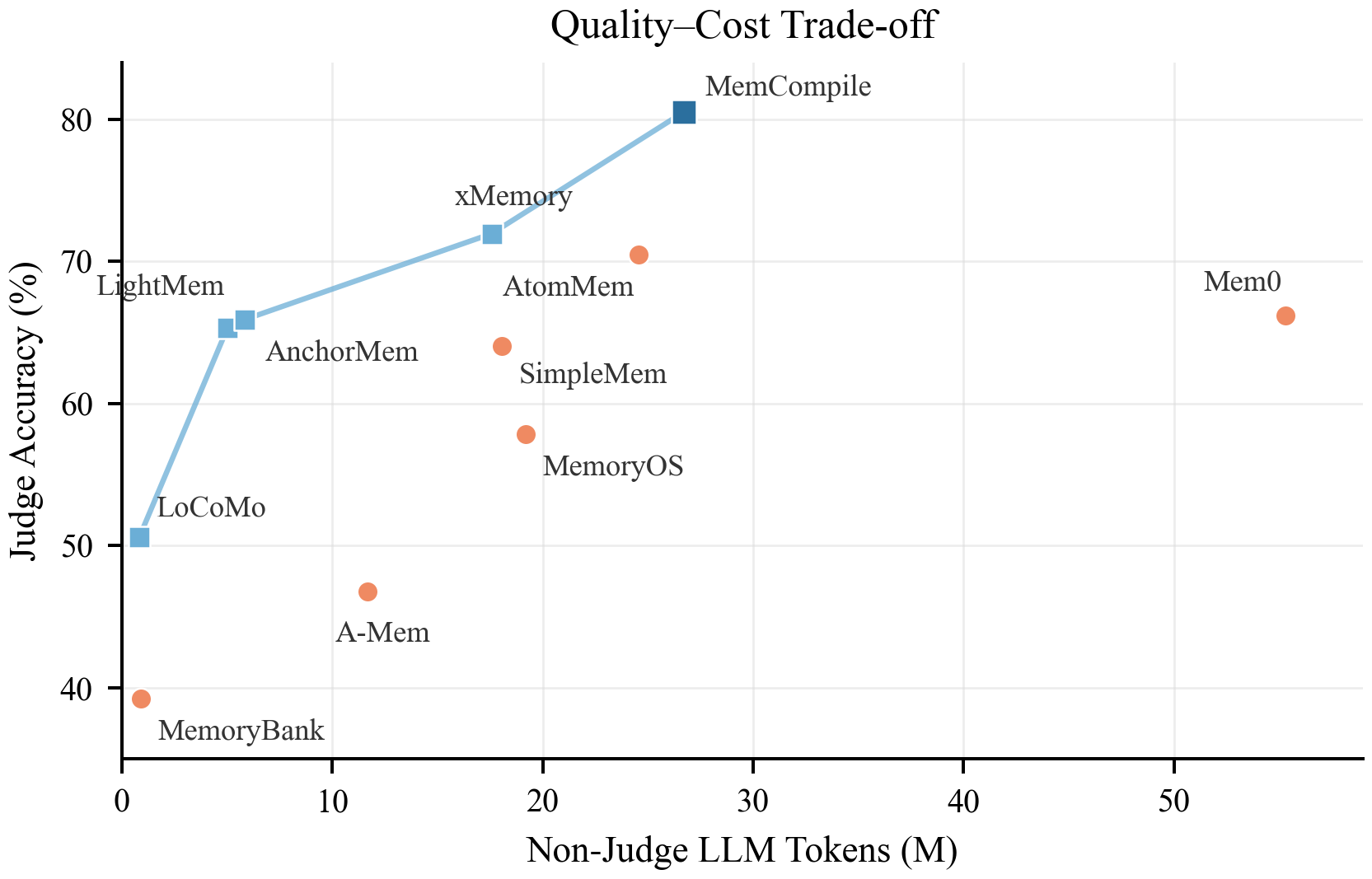}
\caption{\textbf{RQ1 quality-cost comparison}, blue squares form the empirical Pareto frontier.}
\label{fig:pareto}
\end{figure}

\subsection{RQ1: Analysis of End-to-End Quality and Cost}
On LoCoMo, Table~\ref{tab:main} shows that \ours{} ranks first on all reported single-hop, multi-hop, and temporal metrics and obtains the best overall F1/BLEU/Judge scores of \textbf{55.94/48.65/80.52}. Besides the dominant performance on Single-Hop, \ours{} achieves substantial improvements over the strongest baseline on Multi-Hop and Temporal, increasing the J score by \textbf{8.69} and \textbf{7.79} percentage points. This consistent enhancement across direct, compositional, and time-sensitive queries confirms the effectiveness of separating multi-view addressing from source-budgeted evidence compilation: source normalization preserves precise direct recall, while complementary scheduling and packet-scoped rendering remain effective when evidence is distributed across sources or time. 
Figure~\ref{fig:pareto} illustrates the corresponding quality-cost trade-off. Compared with Mem0, \ours{} improves Judge by 14.35 points while using 51.7\% fewer tokens (26.72M vs.\ 55.30M). AtomMem provides the closest cost comparison: \ours{} uses only 8.8\% more tokens (26.72M vs.\ 24.55M), yet improves overall Judge by 10.01 points. Thus, although \ours{} does not minimize absolute token usage, it converts a moderate computational budget into substantially higher answer quality and extends the empirical frontier to 80.52\% Judge. The detailed per phase cost accounting is in Appendix~\ref{app:cost}. 

On LongMemEval-S, \ours{} shows strong cross-benchmark performance. On temporal reasoning and single-session assistant, \ours{ } increases the J score by \textbf{24.1} and \textbf{50} percentage points comparing with Mem0. Table~\ref{tab:longmemeval-results} in Appendix~\ref{app:longmemeval} reports the full results.

\subsection{RQ2: Analysis of Cell Placement}
We first compare five placement layouts that organize the same memory items under a fixed candidate budget. Detailed
layouts definitions are provided in
Appendix~\ref{app:placement-layouts}. Figure~\ref{fig:cell} shows that structured placement improves candidate recovery across complementary query types. Bounded Cell produces the largest gains on single-hop and multi-hop questions, increasing recall by 5.13 percentage points (pp) and 3.42 pp, respectively. It also improves temporal recall by 1.71 pp and open-domain recall by 2.78 pp.



The upstream coverage improvement carries into the Evidence Plane. Removing Cell-routed candidates while keeping the downstream scheduler and evidence compilation pipeline fixed reduces final recall to 81.54\%, a decrease of 3.96 percentage points, F1 to 52.31, BLEU-1 to 45.14, and Judge to 77.53\%. Cell placement expands the set of answer-relevant addresses available to the Evidence Plane, and the downstream pipeline converts this additional source coverage into measurable answer-quality gains.

\begin{table*}[t]
\centering
\small
\setlength{\tabcolsep}{5.0pt}
\begin{tabular}{lrrrrc}
\toprule
Controlled stage ($K=10$) & Unique$\uparrow$ & DBR$\downarrow$ & Gold$\uparrow$ & $\Delta$ Gold vs. previous (95\% CI) & Robust \\
\midrule
Unnormalized claim top-$K$ & 7.690 & 23.08 & 48.89 & --- & --- \\
Exact-text deduplication & 7.742 & 22.56 & 48.56 & $-0.33$ [$-0.87$, 0.21] & --- \\
Source-normalized top-$K$ & \best{9.988} & \best{0.00} & 51.91 & +3.35 [2.45, 4.26] & $\dagger$ \\
$+$ Complementary source scheduler & \best{9.988} & \best{0.00} & \best{80.84} & +28.93 [26.81, 31.14] & $\dagger$ \\
\bottomrule
\end{tabular}
\caption{Controlled source selection at $K=10$. Gold Recall and DBR are percentages. Gold differences are percentage points with paired question-bootstrap 95\% CIs; $\dagger$ means that both question-level and conversation-cluster CIs exclude zero.The source-normalized Top-$K$ ranks packets directly by the maximum claim score.}
\label{tab:causal}
\end{table*}

\subsection{RQ3: Analysis of Source Normalization and Complementary Scheduling}

\paragraph{Multiple hits repeatedly occupy the same source slots}
Table~\ref{tab:source-alias-audit} exposes the mismatch between hit count and source coverage. A Top-20 ranking contains 20 hits but only 14.60 distinct sources on average while same-source hits consume 27.02\% of the nominal budget and affect 97.27\% of questions. Thus adding hits or channels does not translate directly into broader evidence coverage.

\paragraph{source normalization releases duplicate address slot}
Table~\ref{tab:causal} indicates that standard text deduplication provides almost no improvement against directly sending top-10 claims into the reader. While Source normalization instead resolves every hit to its SourcePacket key before applying the budget, yielding a significant but moderate Gold Recall gain of 3.35 pp.

\paragraph{Complementary scheduling converts released capacity into coverage}
Complementary scheduling reallocates the normalized slots toward sources that add missing support. Crucially, the number of distinct sources remains 9.988 and DBR remains zero, while Gold Recall rises from 51.91\% to 80.84\%---a +28.93 pp gain [26.81, 31.14]. Figure~\ref{fig:source-norm} also shows the same large scheduling gain in every category. The contrast separates the two mechanisms: normalization prevents redundant sources from consuming the budget; scheduling determines which complementary sources fill that budget.

\subsection{RQ4: Analysis of Reader Evidence Granularity}
Table~\ref{tab:views} isolates evidence presentation by holding the final SourcePacket exactly constant. Compressing each selected packet to a single representative claim reduces Judge from 80.52\% to 57.01\% indicating that aggressive compression can discard answer-bearing relations, temporal qualifiers, and other context needed to interpret an otherwise correct source.

Presenting the raw SourcePacket recovers much of this loss, reaching 75.32\% Judge, but remains significantly below the complete claim bundle. The compiled bundle preserves the relations and qualifications lost by single-claim compression while filtering the conversational noise retained by raw dialogue. This fixed-source comparison isolates evidence presentation as a separate design problem from source recovery.

\subsection{Ablation Study}

Table~\ref{tab:ablations} compares controlled pipeline paths while
holding the frozen memory and claim-bundle Reader fixed. The full system
achieves the strongest performance across all reported metrics.

A0-A3 separate the effects of anchor construction and subsequent
frontier expansion. These comparisons reveal that frontier expansion compensates most strongly for weaker deterministic anchors, while complementary packet scheduling produces a stronger initial evidence set and reduces reliance on later expansion.

The A4 and A5 variants show an asymmetric complementarity between
BM25 and Cell routing. BM25 provides the stronger standalone route,
whereas adding Cell-derived candidates to BM25 supplies non-redundant
evidence that improves the full pipeline. A6 further suggests that
structured query intent primarily improves evidence prioritization rather
than source coverage while A7 shows that directly retrieving the dense
cosine Top-10 SourcePackets and presenting their claim bundles to the
Reader does not match the compiled pipeline.

\begin{table}[t]
\centering
\small
\setlength{\tabcolsep}{4.3pt}
\begin{tabular}{lrrr}
\toprule
Fixed-packet view & F1$\uparrow$ & BLEU-1$\uparrow$ & Judge$\uparrow$ \\
\midrule
Representative claim & 40.68 & 35.13 & 57.01 \\
Raw SourcePacket & \second{53.00} & \second{46.01} & \second{75.32} \\
Full claim bundle & \best{55.94} & \best{48.65} & \best{80.52} \\
\bottomrule
\end{tabular}
\caption{Reader-view ablation at identical packet Recall (85.50\%).}
\label{tab:views}
\end{table}

\section{Conclusion}
We introduced \ours{}, an Address-Evidence dual-plane architecture that separates memory location from reader-facing compilation. The Address Plane improves addressing and memory placement through organizing source-linked atomic descriptors in bounded Cells. Source normalization then releases duplicate address slot by resolving cross-channel hits to its SourcePacket key; the Evidence Plane schedules complementary sources, preserves initial anchors during bounded expansion, and compiles packet-scoped claims. Experiments on the LoCoMo benchmark confirm that \ours{} achieves state-of-the-art performance while the result on LongMemEval-S demonstrates strong cross-benchmark performance. Ultimately, \ours{} offers a source-normalized and cost-effective evidence-compilation framework for deploying intelligent agents capable of sustaining reliable,
evidence-grounded long-term interactions.

\section*{Limitations}
Evaluation is limited to two conversational-memory benchmarks and proprietary LLMs; broader domains and open-model replications remain future work. Although \ours{} achieves a favorable quality-cost trade-off, its overall token consumption could be further optimized.


\bibliography{custom}

\clearpage

\appendix

\input{appendix}

\end{document}

%% file: appendix.tex
\section{End-to-End Algorithm and Complexity}
\label{app:end-to-end}

\subsection{Serving Pipeline}

Algorithm~\ref{alg:end-to-end} specifies the complete serving path. The
construction stage creates one immutable SourcePacket per dialogue turn,
extracts an ordered set of atomic descriptors, records a foreign key from each
descriptor to its packet, and places the descriptors into bounded Cells. At
query time, query-only intent fields augment the lexical query. Cell routing
and BM25 produce atomic candidates, which are collapsed by SourcePacket
identity before listwise scheduling. The semantic frontier operates over a
separate packet index and can only append packets to the scheduled anchors.
Finally, the reader view is materialized without changing the selected packet
sequence.

\begin{algorithm*}[t]
\caption{End-to-end query-conditioned memory compilation}
\label{alg:end-to-end}
\begin{algorithmic}[1]
\REQUIRE Ordered dialogue turns $\mathcal H$, query $q$
\ENSURE Reader context $X_q$
\STATE $\mathcal P\leftarrow\emptyset$, $\mathcal X\leftarrow\emptyset$,
$\mathcal C\leftarrow\emptyset$
\FOR{each turn $d$ in dialogue order}
  \STATE register immutable packet $p_d$ with content hash
  \STATE extract standalone facts $F_d$ using $d$ and its preceding turn
  \FOR{each retained fact $f\in F_d$ in extraction order}
    \STATE extract people, keywords, event time, and interaction time
    \STATE construct descriptor $x$ and set $\pi(x)\leftarrow p_d$
    \STATE place $x$ by Algorithm~\ref{alg:cell-placement}
  \ENDFOR
\ENDFOR
\STATE extract query-only intent $I_q$ and form normalized query terms
\STATE $\mathcal K_q\leftarrow$ Cell/BM25 candidates from
Algorithm~\ref{alg:cell-routing}
\STATE $\mathcal U_q\leftarrow\operatorname{CollapseBySource}(\mathcal K_q)$
\STATE $\mathcal A_0\leftarrow
\operatorname{ListwiseSchedule}_{10}(q,\mathcal U_q)$
\STATE $\mathcal D_q\leftarrow\operatorname{SemanticTop}_{40}(q^+,\mathcal P)$
\STATE $\sigma_q\leftarrow
\operatorname{ListwiseSchedule}(q,\mathcal A_0\cup\mathcal D_q)$
\STATE $\Delta_q\leftarrow$ first two packets in
$\sigma_q[1{:}5]\setminus\mathcal A_0$
\STATE $\mathcal E_q\leftarrow\mathcal A_0\mathbin{\Vert}\Delta_q$
\FOR{each $p\in\mathcal E_q$ in scheduled order}
  \STATE $B_p\leftarrow\langle x\in\mathcal X:\pi(x)=p\rangle$
  \STATE append packet timestamp and ordered claim bundle $B_p$ to $X_q$
\ENDFOR
\RETURN $X_q$
\end{algorithmic}
\end{algorithm*}

\subsection{Source Normalization and Scheduler Validation}
\label{app:scheduler}

Atomic candidates are grouped by their stored \texttt{source\_packet\_id}.
Within each group, the descriptor with the largest retrieval score becomes the
representative address; ties retain the earliest candidate. Packet ordering is
then descending by the representative score, followed by first candidate
position and packet identifier. This operation changes the selection state
from claims to sources without adding any source absent from the closed
candidate pool.

For listwise scheduling, candidate packets are assigned opaque labels
\texttt{C001}, \texttt{C002}, and so forth. The request displays each label
with the packet's dialogue identifier, speaker, recorded date, and content.
The requested cardinality is $\min(K,|\mathcal U_q|)$. A valid response must
contain exactly that many distinct labels, all drawn from the request. Invalid
responses are retried up to three times with a schema correction; after the
final attempt, the parser retains only valid, unique labels in response order.
If fewer than $K$ labels survive, the remaining closed candidate pool is
submitted in a subsequent round until the target cardinality is reached. Cache
identity binds the query, ordered candidate tensor, rendered packet contents,
prompt, model, protocol version, and $K$.

\subsection{Complexity}

Let $N=|\mathcal P|$ be the number of source packets, $M=|\mathcal X|$ the
number of atomic descriptors, $C=|\mathcal C|$ the number of Cells, $M_C$ the
item capacity per Cell, and $d$ the embedding dimension. With bounded $M_C$,
online Cell placement costs $O(MCM_C)$ in the direct implementation. The
lexical query pass scores $M$ descriptor documents,
while Cell routing scores $C$ Cells and examines at most six opened Cells plus
the bounded BM25 pool. Dense packet retrieval costs $O(Nd)$ for a matrix--vector
product and $O(N\log 40)$ for a comparison-based Top-40 implementation. Source
normalization is linear in the atomic candidate count. Reader compilation is
linear in the number of claims attached to the selected packets. LLM calls for
fact extraction, metadata extraction, the two listwise schedules, and answer
generation dominate wall-clock cost and are accounted for separately in
Appendix~\ref{app:cost}.

\subsection{Source-Aliasing Audit}

Eligible $N$ is the number of questions for which the retrieval stage returned at least the requested depth. DBR is the fraction of retrieved slots beyond the first occurrence of each distinct SourcePacket. A question is Affected if at least one retrieved slot duplicates a SourcePacket already present at that depth.

\begin{table}[t]
\centering
\scriptsize
\setlength{\tabcolsep}{2.5pt}
\resizebox{\columnwidth}{!}{%
\begin{tabular}{lrrrrr}
\toprule
Depth & Eligible $N$ & Unique & DBR$\downarrow$ & Affected & Max-source $\mu$/P95/max \\
\midrule
Top-10 & 1,539 & 7.69 & 23.08\% & 84.21\% & 2.42/4/6 \\
Top-20 & 1,537 & 14.60 & 27.02\% & 97.27\% & 3.18/5/7 \\
Top-40 & 1,533 & 26.94 & 32.66\% & 100.00\% & 4.12/6/8 \\
\bottomrule
\end{tabular}}
\caption{Pre-normalization aliasing at three retrieval depths. DBR and
Affected are percentages. Max-source reports the mean, 95th percentile, and
global maximum number of slots occupied by one source.}
\label{tab:source-alias-audit}
\end{table}

\begin{figure*}[t]
\centering
\includegraphics[width=0.78\textwidth]{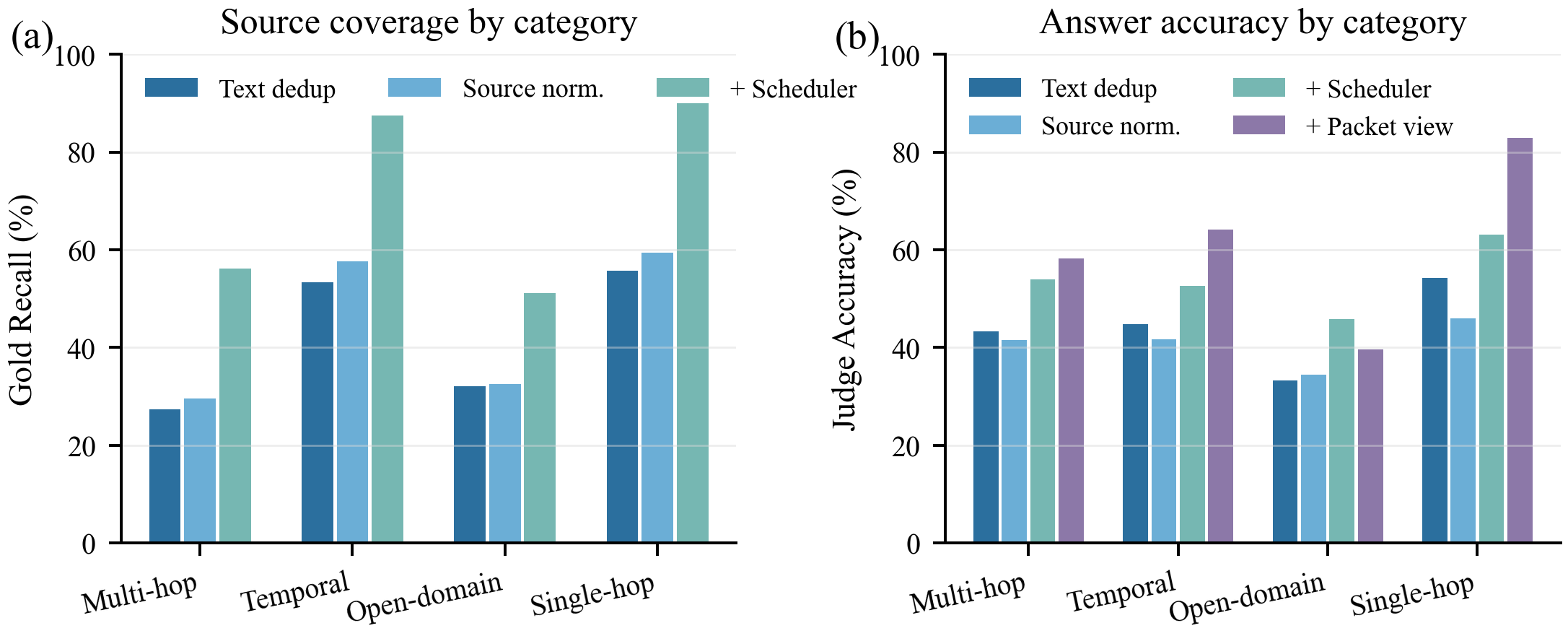}
\caption{Category-level effects of source normalization, complementary
scheduling, and source-scoped claim-bundle compilation. (a) Gold Source Recall
after exact-text deduplication, source-ID normalization, and complementary
source scheduling. (b) Judge Accuracy across the controlled stages and the
fixed-source claim-bundle reader view.}
\label{fig:source-norm}
\end{figure*}

\section{Descriptor Extraction and Cell Construction}
\label{app:cell}

\subsection{Source-Linked Descriptor Construction}

For a turn $d_i$, fact extraction observes the current turn, its immediately
preceding turn, and the recorded session time. Greetings, acknowledgements,
system-context statements, and configured low-value prefixes are removed.
Each retained fact receives an identifier composed of the conversation,
dialogue-turn, and within-turn fact indices. Metadata extraction supplies up
to five keyword phrases, involved people, an event time when supported, the
interaction time, and event-role fields. The main experimental layout is the
bounded lexical Cell layout: its Cell topology uses descriptor text, people,
keywords, session index, and token count. Each descriptor is linked to exactly
one immutable packet, and packet claim identifiers retain construction order.

\subsection{Bounded Cell Placement}

For descriptor $x_i$ and Cell $c$, define keyword and person Jaccard
similarities $J_K$ and $J_P$. The temporal term is

\begin{equation}
T(i,c)=
\begin{cases}
1, & j_i\in[j_c^{\min},j_c^{\max}+1],\\
0.25, & \text{otherwise}.
\end{cases}
\end{equation}

The implemented coherence is
\begin{equation}
\begin{aligned}
C(i,c)={}&0.45J_K(i,c)+0.35J_P(i,c)\\
&+0.20T(i,c).
\end{aligned}
\label{eq:appendix-cell-coherence}
\end{equation}
The locality gain is
\begin{equation}
\begin{aligned}
L(i,c)=\log\!\bigl(1&+|\mathsf{Key}_i\cap\mathsf{Key}_c|\\
&+|\mathsf{Per}_i\cap\mathsf{Per}_c|\bigr).
\end{aligned}
\end{equation}
The inconsistency penalty $X(i,c)$ is $0.45$ when an existing descriptor in
$c$ shares a person and at least one keyword with $x_i$ but only one of the two
contains a negation marker; otherwise it is zero. Let $H(c)$ be the sum of
keyword entropy, person entropy, a log-scaled session-span penalty, and the
fraction of repeated dialogue identifiers:
\begin{align}
H(c)={}&H_K(c)+H_P(c)\nonumber\\
&+0.2\log(1+j_c^{\max}-j_c^{\min})\nonumber\\
&+\frac{\sum_d\max(0,n_d-1)}{|c|}.
\end{align}
Writing $r_c=\max(1,B_C-\operatorname{Tok}(c))$ and
$\Delta_H=\max(0,H(c\cup\{x_i\})-H(c))$, expansion pressure is
\begin{align}
E(i,c)={}&\frac{n_i}{r_c}+0.4\Delta_H\nonumber\\
&+0.25\max\!\left(0,\frac{|c|+1}{M_C}-C(i,c)\right)\nonumber\\
&+0.25\max(0,H(c)-1.8).
\label{eq:appendix-expansion}
\end{align}
The final placement score is
$L(i,c)+1.4C(i,c)-X(i,c)-0.45E(i,c)$. A new Cell is scored with
$L=0$, $C=0.55$, $X=0$, and $E=0.65+n_i/B_C$. The experiment uses
$B_C=360$ and $M_C=12$.

\begin{algorithm*}[t]
\caption{Online bounded Cell placement}
\label{alg:cell-placement}
\begin{algorithmic}[1]
\REQUIRE Descriptor $x_i$, current Cells $\mathcal C$
\STATE $Q\leftarrow\emptyset$
\FOR{each $c\in\mathcal C$}
  \IF{$\operatorname{Tok}(c)+n_i\leq360$ and $|c|<12$}
    \STATE compute $L(i,c),C(i,c),X(i,c),E(i,c)$
    \STATE append $(L+1.4C-X-0.45E,c)$ to $Q$
  \ENDIF
\ENDFOR
\STATE create candidate $c_{\mathrm{new}}$ with score
$1.4(0.55)-0.45(0.65+n_i/360)$
\STATE append $(S_{\mathrm{new}},c_{\mathrm{new}})$ to $Q$
\STATE $c^*\leftarrow\arg\max_{(S,c)\in Q}S$
\STATE append $x_i$ to $c^*$ and recompute token, people, keyword, session,
title, and entropy summaries
\end{algorithmic}
\end{algorithm*}

\subsection{Cell Routing and Candidate Union}
\label{app:cell-routing}
 BM25 scores all descriptors and retains
the top 96; its first eight descriptors seed Cell scoring. For query term set
$Q$, person set $P$, and Cell $c$, the base relevance is
\begin{align}
u(c,q)={}&2J(Q,K_c)+2.2J(P,P_c)\nonumber\\
&+0.7J(Q,\operatorname{Tok}(t_c))+T_q(c)\nonumber\\
&+0.03\frac{\log(1+|c|)}{\log(1+M_C)},
\end{align}
where $T_q(c)=0.25$ for a temporal query when $c$ contains temporal metadata
and is zero otherwise. Cells with base relevance below $0.08$ are excluded.
The route pool contains the first
$\max(12,6B_{\mathrm{cell}})=36$ Cells, with $B_{\mathrm{cell}}=6$.

At step $t$, redundancy is the Jaccard overlap between the candidate Cell's
people/keywords and their union over already opened Cells. The implemented
marginal gain is
\begin{align}
g(c\mid\mathcal O_t,q)={}&u(c,q)(1-0.45R(c,\mathcal O_t))\nonumber\\
&-0.02(1+|\mathcal O_t|)\nonumber\\
&-0.0005\operatorname{Tok}(c).
\end{align}
After at least one Cell is opened, routing stops when $g<0$ or the
redundancy-adjusted utility falls below $0.68$ of the best Cell utility. It
also stops once cumulative adjusted utility reaches $90\%$ of positive route
pool utility, or after six Cells. Descriptors from opened Cells are united with
the BM25 Top-96. Descriptors reached by both channels receive the configured
BM25 contribution, and the complete candidate tensor records both channel
identities before source normalization.

Descriptor relevance combines text, keyword, person, and temporal matches as
specified in Eq.~\ref{eq:claim-score}, with
$(\alpha_f,\alpha_k,\alpha_p,\alpha_t,\alpha_b)
=(2.0,1.6,2.4,0.3,0.45)$. BM25 uses $k_1=1.5$ and $b=0.75$ over
documents formed by descriptor text, keywords, and people. Raw BM25 values are
mapped to $2\log(1+s_i)/\log(1+s_{\max})$ before applying the weight $0.45$.
For the diagnostic pre-scheduler Top-10, descriptors are first consolidated by
dialogue identifier. The representative relevance is augmented by
$0.10\log(1+\sum_i\max(0,r_i))$ over same-source support. Greedy MMR then
subtracts $0.58$ times maximum token-set overlap with an already selected
representative and $0.08$ for each already selected representative from the
same Cell. The query-conditioned packet scheduler instead receives the complete
source-normalized candidate tensor, so this diagnostic Top-10 does not restrict
its closed candidate pool.

\begin{algorithm*}[t]
\caption{Diminishing-benefit Cell routing}
\label{alg:cell-routing}
\begin{algorithmic}[1]
\REQUIRE Query $q$, Cells $\mathcal C$, descriptors $\mathcal X$
\STATE extract query-only people, keywords, and time; normalize $Q,P$
\STATE $S\leftarrow\operatorname{BM25Top}_{96}(Q,\mathcal X)$; seed with $S[1{:}8]$
\STATE score Cells by $u(c,q)$ and discard base score below $0.08$
\STATE $V\leftarrow$ first 36 scored Cells; $\mathcal O\leftarrow\emptyset$
\FOR{each $c\in V$ in descending $u(c,q)$}
  \STATE compute redundancy-adjusted utility and marginal gain $g$
  \IF{at least one Cell is open and $g<0$ or relative utility $<0.68$}
    \STATE \textbf{break}
  \ENDIF
  \STATE append $c$ to $\mathcal O$
  \IF{$|\mathcal O|=6$ or cumulative adjusted utility reaches $90\%$}
    \STATE \textbf{break}
  \ENDIF
\ENDFOR
\RETURN descriptors in $\mathcal O$ united with $S$
\end{algorithmic}
\end{algorithm*}

\subsection{Controlled Placement Layouts}
\label{app:placement-layouts}

\paragraph{Controlled intervention}
All layouts use the same atomic memory items, BM25 baseline ranking, total candidate budget, question set, source packet scheduler, semantic frontier, Reader, and Judge. Evaluation is read-only with respect to persistent memory. 

\paragraph{Flat}
Flat disables Cell-derived candidates rather than placing all items in one container. Its candidate sequence is the BM25-only sequence used by the no-Cell endpoint. Flat establishes the candidate recall of 0.8154 against which every structured layout is measured.

\paragraph{Bounded Cells}
For every new item, all capacity-feasible existing Cells compete with a new Cell under the placement objective in Eq.~\ref{eq:cell-placement-score}. The resulting organization jointly reflects person and keyword locality, session proximity, coherence, conflict avoidance, capacity, and entropy pressure. The policy is deterministic and accumulative, not a gradient-trained clustering model. The bouned cell layout improves recall by 3.959 points.

\paragraph{Sequential}
Sequential orders items by session index and then by dialogue identifier within each session order, and partitions the resulting sequence into fixed-length contiguous blocks. It discards people, keywords, coherence, and conflict signals, retaining only dialogue adjacency. This layout tests whether simple sequential locality accounts for the Cell benefit.

\paragraph{Session}
Session first groups items by session index, preserves their within-session order, and partitions each session independently into fixed-length blocks. Comparing Session with Sequential therefore isolates the value of the dataset's session boundaries. 

\paragraph{Entity}
Entity assigns each item to a bucket keyed by its primary person entity, orders each bucket by dialogue order, and partitions it into fixed-length blocks. It tests person identity as a cross-turn and cross-session address without using the complete placement objective. Entity layout improves the recall by 2.910 points. The result identifies person locality as a useful but incomplete component of the persistent organization.

\begin{figure*}[t]
\centering
\includegraphics[width=0.78\textwidth]{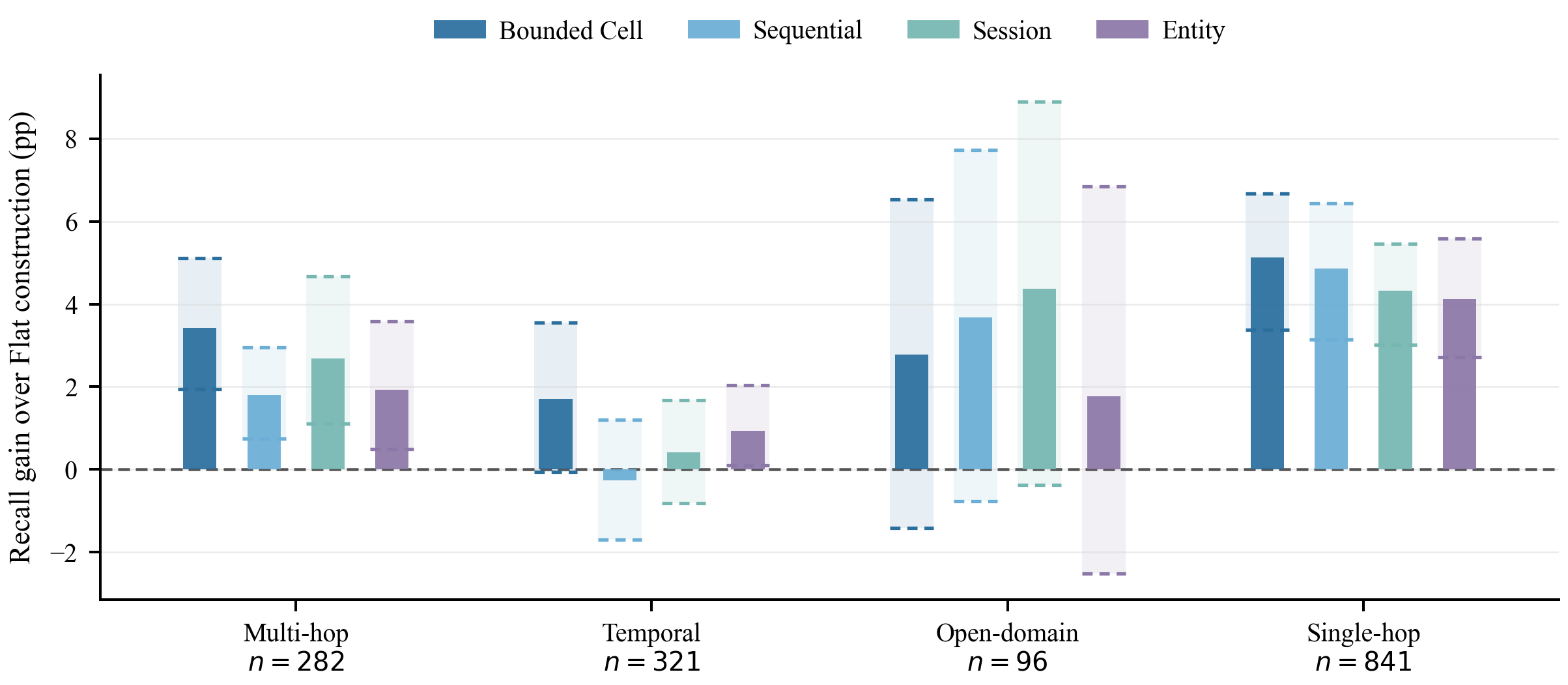}
\caption{Cell-placement contribution to recall. Bars are
gains over Flat construction; translucent bands and dashed endpoints show
conversation-bootstrap 95\% intervals.}
\label{fig:cell}
\end{figure*}

\section{Full Experimental Configuration}
\label{app:configuration}

\subsection{LoCoMo Protocol}

The LoCoMo evaluation contains 1,540 questions from ten conversations and retains the benchmark's single-hop, multi-hop, temporal, and open-domain categories. Memory construction receives conversation sessions only. Query text is exposed to query-intent extraction and retrieval at evaluation time; answers, question categories, gold evidence, and judge labels are not inputs to construction, addressing, scheduling,  frontier selection and answer generation. Evaluation is read-only with respect to persistent memory state.

\begin{table*}[t]
\centering
\small
\begin{tabular}{lll}
\toprule
Component & Setting & Value \\
\midrule
Models & construction, intent, scheduling, answer & \texttt{gpt-4o-mini} \\
 & dense retrieval & \texttt{text-embedding-3-small} \\
 & temperature / maximum completion & $0$ / 2,048 tokens \\
\midrule
Cell construction & layout / token cap / item cap & bounded lexical Cell / 360 / 12 \\
 & $\beta,\gamma,\eta$ / new-Cell penalty & 1.4, 1.0, 0.45 / 0.65 \\
 & entropy weight / entropy threshold & 0.4 / 1.8 \\
\midrule
Addressing & maximum/minimum opened Cells & 6 / 1 \\
 & admission / open / token costs & 0.08 / 0.02 / 0.0005 \\
 & relative floor / cumulative top-$p$ & 0.68 / 0.90 \\
 & BM25 seeds / BM25 candidates & 8 / 96 \\
 & BM25 weight / descriptor support weight & 0.45 / 0.10 \\
 & descriptor MMR duplicate / same-Cell weights & 0.58 / 0.08 \\
\midrule
Evidence Plane & initial SourcePackets $K_0$ & 10 \\
 & semantic packet proposals & 40 \\
 & frontier rank cutoff / append cap & 5 / 2 \\
 & reader view & source-scoped atomic-claim bundle \\
\midrule
Evaluation & answer metrics & token F1, BLEU-1, Judge Accuracy \\
 & judge model & \texttt{gpt-5-nano} \\
 & source metric & Gold Source Recall \\
 & generation cost metric & construction + query-time LLM tokens \\
\bottomrule
\end{tabular}
\caption{Configuration of the reported MemCompile experiment.}
\label{tab:full-config}
\end{table*}

\subsection{Dense Frontier}
Each packet document concatenates raw turn text, image caption and query fields
when present, all retained packet claims, packet-level people and keywords, and
the recorded session time. Documents and the enriched query are embedded and
$\ell_2$-normalized; cosine similarity is therefore a matrix--vector dot
product. The enriched query concatenates the question with query-only people,
keywords, and time. The semantic Top-40 is united with the ten scheduled
anchors by dialogue identifier. A second closed-pool listwise schedule orders
this union. All ten anchors are preserved, and at most two previously
unselected packets among the first five scheduled positions are appended.

\subsubsection{Frontier Hyperparameter Sensitivity}
\label{app:frontier-sensitivity}

\begin{figure*}[t]
\centering
\includegraphics[width=0.78\textwidth]{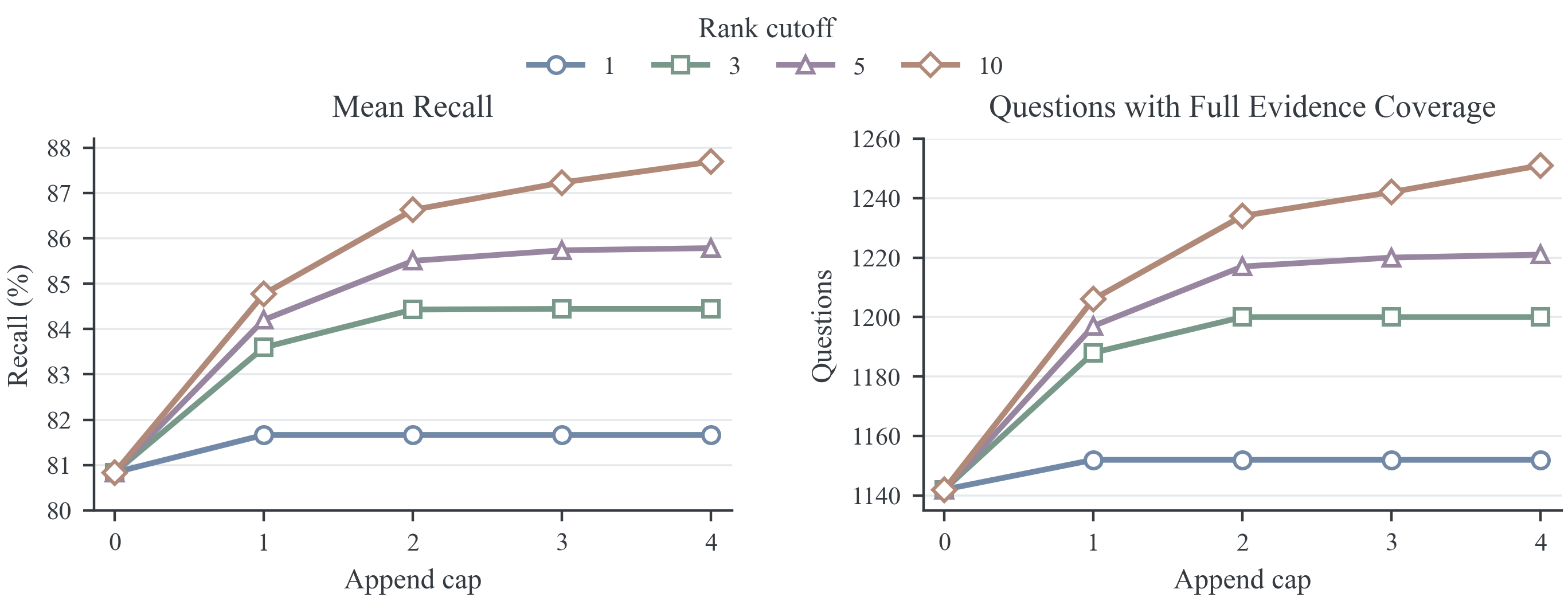}
\caption{Post-hoc frontier sensitivity around the fixed reported configuration.}
\label{fig:frontier-sensitivity}
\end{figure*}

We analyze how the frontier budgets affect evidence coverage on the
full LoCoMo benchmark, with $K_D=40$ fixed. This diagnostic sweep
characterizes the behavior of the bounded frontier around the reported
configuration.

Figure~\ref{fig:frontier-sensitivity} shows the interaction between the
rank cutoff $K_R$ and append cap $K_\Delta$. When $K_\Delta=0$, all
curves coincide because no frontier proposal can enter the final evidence
set. For $K_R=1$, increasing the cap beyond one has no effect by
construction. Larger rank windows make more frontier proposals eligible,
but for $K_R\in\{3,5\}$ most of the observed coverage gain is obtained
by $K_\Delta=2$. In contrast, $K_R=10$ continues to benefit from larger
append caps, indicating that a wider rank window is useful mainly when
more context growth is permitted.

At the fixed setting $(K_R,K_\Delta)=(5,2)$, mean Gold Source Recall
reaches 85.50\%, with 1,217 questions achieving complete gold-source
coverage. Increasing the append cap from two to four adds only 0.28
recall points and four fully covered questions. Because $K_\Delta=2$
limits expansion to at most two packets beyond the ten scheduled anchors,
this setting represents a bounded coverage--context operating point  capturing most of the observed gain for $K_R\leq5$ while limiting
additional reader context.

\subsection{Reader Views}

The representative-claim view renders the highest-scoring address claim for
each selected source. The raw-source view renders the packet's text and, when
present, its image caption and query fields. The claim-bundle view renders all
retained claims linked to the selected packet in construction order. Packet
selection and ordering are held fixed across the three views.

\section{LongMemEval Results}
\label{app:longmemeval}

LongMemEval examples are converted into the same neutral session/turn schema used by LoCoMo. Each official haystack session retains its identifier and date; each turn receives a stable identifier formed from the session ID and turn position. The system constructs a separate memory for each evaluation instance and applies the same source-linked descriptor, Cell routing, SourcePacket scheduling, semantic frontier, and reader compilation pipeline. The official question types comprise single-session user, single-session
assistant, single-session preference, temporal reasoning, knowledge update, and multi-session questions. Answer-session and turn-level evidence labels are used only for offline evidence evaluation. On LongMemEval-S, \ours{} achieves the best Judge Accuracy among
the evaluated methods in five of the six reported categories: single-session user (91.43\%), temporal reasoning (72.18\%), knowledge update (78.21\%), single-session assistant (94.64\%), and single-session preference (70.00\%). Table~\ref{tab:longmemeval-results} reports the full category-level results.

\begin{table*}[t]
\centering
\small
\setlength{\tabcolsep}{5pt}
\begin{tabular}{lrrr}
\toprule
Category & \ours{} & LightMem~\cite{fang2026lightmem} & Mem0~\cite{mem0} \\
& J$\uparrow$ & J$\uparrow$ & J$\uparrow$ \\
\midrule
SSU & \best{91.43} & 84.29 & 90.00 \\
MS  & 54.14 & 57.14 & \best{65.41} \\
TR  & \best{72.18} & 51.88 & 48.12 \\
KU  & \best{78.21} & 76.92 & 74.36 \\
SSA & \best{94.64} & 33.93 & 44.64 \\
SSP & \best{70.00} & 60.00 & 63.33 \\
\bottomrule
\end{tabular}
\caption{Judge accuracy (\%) on LongMemEval-S. Best category-level results are bold. SSU, MS, TR, KU, SSA, and SSP denote single-session user, multi-session, temporal reasoning, knowledge update, single-session assistant, and single-session preference, respectively.}
\label{tab:longmemeval-results}
\end{table*}

\begin{table*}[t]
\centering
\small
\setlength{\tabcolsep}{12pt}
\begin{tabular}{lrrr}
\toprule
Stage & Total tokens & Tokens/question & Share of non-Judge total \\
\midrule
Fact extraction & 1,578,484 & 1,025 & 5.9\% \\
Metadata extraction & 12,492,607 & 8,112 & 46.8\% \\
\textbf{Cold-build subtotal} & \textbf{14,071,091} & \textbf{9,137} & \textbf{52.7\%} \\
\midrule
Query intent & 313,178 & 203 & 1.2\% \\
packet scheduler & 6,170,000 & 4,006 & 23.1\% \\
frontier reranker & 4,380,000 & 2,844 & 16.4\% \\
Answer generation & 1,786,551 & 1,160 & 6.7\% \\
\textbf{Online subtotal} & \textbf{12,649,729} & \textbf{8,214} & \textbf{47.3\%} \\
\midrule
\textbf{Non-Judge end-to-end} & \textbf{26,720,820} & \textbf{17,351} & \textbf{100.0\%} \\
\bottomrule
\end{tabular}
\caption{Generation-token accounting for \ours{} on the 1,540 LoCoMo
questions. Totals include prompt and completion tokens from unique successful
LLM calls; Judge and embedding usage are excluded. Per-question costs amortize
each total over all 1,540 questions, and shares are rounded independently.}
\label{tab:cost-breakdown}
\end{table*}

\section{Cost Accounting}
\label{app:cost}

The reported generation-token cost is the sum of prompt and completion tokens
for unique successful LLM requests used in memory construction and query-time
serving:
\begin{align}
T_{\mathrm{reported}}={}&T_{\mathrm{fact}}+T_{\mathrm{metadata}}
+T_{\mathrm{intent}}\nonumber\\
&+T_{\mathrm{schedule}}+T_{\mathrm{frontier}}
+T_{\mathrm{answer}}.
\end{align}
Judge calls and embedding tokens are recorded separately and are excluded from
the Tokens column in Table~\ref{tab:main}. Construction cost is incurred once
per conversation. Query-time cost includes query-intent extraction, packet
scheduling, frontier scheduling, and answer generation. Reader evidence length
is recorded separately from API-reported LLM usage.

Successful LLM responses are stored under an identity containing the system
prompt, user request, response format, model, temperature, maximum completion
length, and call type. A cache hit contributes zero incremental tokens to the
current invocation. Durable usage is reconstructed from unique successful
cache entries, preventing both double counting after replay and omission when
an API response is committed before the coordinating checkpoint. Embedding
requests use a separate identity-addressed cache and an idempotent usage
ledger. Packet scheduler caches additionally bind the ordered candidate tensor
and rendered packet content hashes.

Cold construction accounts for 14.07M tokens, or 52.7\% of non-Judge usage,
and is amortized across subsequent questions over the same conversation.
Online serving accounts for the remaining 12.65M tokens. Within serving, the
R2 packet scheduler and E0.7 frontier reranker consume 10.55M tokens in total,
equivalent to 39.5\% of the end-to-end total (and 83.4\% of online usage).
These two stages are therefore the principal online cost paid for packet-level
selection and progressive evidence compilation; query-intent extraction and
answer generation together account for the remaining 7.9\% of the total.

\section{Prompts and Output Parsing}
\label{app:prompts}

The following boxes reproduce the stored prompt artifacts verbatim. All baselines use the same prompt settings as \ours{} to ensure a fair comparison.

\begin{promptbox}[prm:metadata]{Fact Metadata Extraction}
Extract factual metadata as one JSON object. Use only information supported by the standalone Fact and supplied dialogue context.

Return:
- `people`: complete person names explicitly involved.
- `keywords`: at most five concise noun phrases useful for lexical retrieval; exclude people and time.
- `time`: `[event_time, interaction_time]`; normalize an explicit/inferable event date, otherwise use `""`; interaction time is the supplied session time.
- `needs_profile_extraction`: true only for a stable personal attribute.
- `event`: one event/state represented by:
  - `subject_people`: subjects of the action/state, not all mentioned participants.
  - `predicate`: one lowercase English base-form verb lemma; never include auxiliaries or arguments.
  - `focus_terms`: one or more short noun/state phrases directly changed, chosen, evaluated, owned, experienced, or described by this Fact. Prefer wording grounded in the Fact.
  - `goal_context_terms`: optional wider named endeavor, purpose, task, or background explicitly supported by the Fact and potentially shared with distinct events. It is not the predicate/action, an emotion or quality, a time expression, or a generic category. Leave it empty unless the Fact states a genuinely wider context.
  - `event_type`: `occurrence`, `plan`, `state`, `preference`, or `relationship`.
  - `event_time`: same normalized event time.
  - `polarity`: `positive`, `negative`, `neutral`, or `unknown`.
  - `confidence`: 0.0 to 1.0.

Distinct direct focuses remain distinct events even when they share a person, predicate, or goal. A repeated or elaborating statement about the same direct focus may represent the same event. Use empty arrays rather than invented content.

JSON shape:
{"people":[],"keywords":[],"time":["", ""],"needs_profile_extraction":false,"event":{"subject_people":[],"predicate":"","focus_terms":[],"goal_context_terms":[],"event_type":"occurrence","event_time":"","polarity":"unknown","confidence":0.0}}

\end{promptbox}

\begin{promptbox}[prm:query-intent]{Query Intent Extraction}
Extract query-only retrieval intent as one JSON object. Do not answer the
query, use conversation memory, or infer facts not stated or implied by the
query.

Return:
- `need_attribute`: whether answering requires a stable personal attribute,
  preference, relationship, occupation, health fact, or long-term goal;
- `people`: complete person names explicitly mentioned in the query;
- `keywords`: at most five concise topics, concepts, or named entities useful
  for retrieval, preferably normalized noun phrases;
- `time`: explicit absolute or relative time constraints from the query, or
  null when the query supplies none.

JSON shape:
{"need_attribute":false,"people":[],"keywords":[],"time":null}
\end{promptbox}

\begin{promptbox}[prm:packet-scheduler]{Initial SourcePacket Scheduler}
You rank conversation source packets for evidence retrieval.

Given a query and a closed candidate list, return exactly the requested number
of distinct candidate labels in descending usefulness for answering the query.
Rank packets that directly contain answer-bearing evidence first. When the
answer requires multiple facts, include complementary packets that jointly
support it. Do not answer the query and do not invent labels.

Return one JSON object with this schema:
{"ranked_candidate_ids": ["C001", "C002"]}

Every returned string must be one of the bracketed `C###` labels in the
candidate list. Never return a `dia_id`, packet ID, explanation, duplicate,
or label that is absent from the candidate list. Count the distinct labels
before returning the object.
\end{promptbox}

\begin{promptbox}[prm:frontier]{Evidence Frontier Reranker}
You are a query-conditioned evidence selector.

Given a question and a closed list of source packets, return exactly the
requested number of distinct candidate labels, ordered from most to least
useful for answering the question.

Select packets that directly support the requested answer slot. Preserve
complementary packets when the answer needs a relation, a temporal anchor and
event, a comparison, or multiple list members. Prefer explicit evidence over
mere topical or name overlap. Avoid redundant packets when a different packet
adds necessary answer-bearing information. Do not answer the question and do
not infer facts that are absent from the candidates.

Return one JSON object with this schema:
{"ranked_candidate_ids": ["C001", "C002"]}

Every string must be one of the bracketed C### labels in the candidate list.
Never return a dia_id, packet ID, explanation, duplicate, or absent label.
Count the distinct labels before returning the object.
\end{promptbox}

\begin{promptbox}[prm:answer-direct]{Answer Generation: Direct and Compositional}
Answer from the selected evidence only.

First identify the answer slot requested by the question: person, place,
object, number, date, duration, reason, feeling, yes/no judgment, or list. Then
output only the minimum text that fully answers that slot.

Rules:
- Reuse the most direct and specific wording from the evidence whenever possible.
- For one value, give one short phrase rather than a sentence.
- For a list, merge all and only the explicitly supported requested items into
  a concise comma-separated list. Do not drop supported items merely to shorten it.
- For yes/no or likelihood questions, begin with Yes or No and add only a
  necessary short qualifier supported by the evidence.
- Directly perform evidence-supported merging, counting, comparison, and simple
  time conversion. These operations are not reasons to abstain.
- Match the question's date or duration granularity. Resolve relative time only
  from a reliable date anchor in the evidence; otherwise retain the supported
  expression.
- Preserve negation, subject, and time qualifiers needed for correctness.

Do not repeat the question, explain, cite sources, add politeness, use outside
knowledge, or infer unsupported personal facts. Output "Unknown" only if all
selected evidence is insufficient.
\end{promptbox}

\begin{promptbox}[prm:answer-temporal]{Answer Generation: Temporal}
You are answering a question that requires temporal reasoning over selected
conversation evidence.

First identify the final answer slot. A date in the question may only constrain
which event is relevant. If the question asks for a person, place, object,
activity, or other non-time value, use the timeline to select the event and
output only that value. Apply temporal output formatting only when the final
requested slot is a time or duration.

Temporal expression rules:
1. Use only the selected evidence.
2. If a clear absolute event date is directly supported and there is no
   conflicting anchored relative expression, use the absolute date.
3. Resolve simple offsets such as last year or last month from a reliable
   event-linked anchor.
4. For week- or weekday-based expressions such as last week, next Saturday, or
   last Sunday, do not calculate an exact date. Preserve the relative expression
   or combine it with its reliable anchor, such as "the week before 9 July 2023"
   or "the Sunday before 22 May 2023".
5. If only a month is supported, return the month and year. If only a year is
   supported, return the year.
6. For duration, return only the shortest duration value without a "Duration:"
   label. Perform evidence-supported counting and comparison.
7. Do not treat a record date by itself as an event date. Preserve event,
   subject, negation, and ownership.

Output only the shortest final value for the requested slot. Do not include
reasoning, explanation, citation, or sentence framing. Output Unknown only if
all selected evidence is insufficient.
\end{promptbox}

\begin{promptbox}[prm:judge]{LLM Judge}
Your task is to label an answer to a question as "CORRECT" or "WRONG".

You will be given:
1. a question,
2. the gold answer,
3. a generated answer.

The question asks about something one user should know about another user based
on prior conversations. Be generous: if the generated answer refers to the same
topic or same fact as the gold answer, mark it CORRECT even if the wording is
longer, shorter, or formatted differently.

For time-related questions, the gold answer may be a date, month, year, or
relative time expression. Mark the generated answer CORRECT if it refers to the
same time period, even if the format differs.

Question: {question}
Gold answer: {gold_answer}
Generated answer: {generated_answer}

Return only JSON with this schema:
{{"label": "CORRECT"}}

The label must be exactly "CORRECT" or "WRONG".
\end{promptbox}

\subsection{Dynamically Constructed Requests}

Fact extraction uses the following system message exactly:
Its user request supplies the preceding turn when present, followed by the
current dialogue identifier, speaker, session time, text, and available image
caption or image keywords. Metadata extraction supplies the standalone fact,
preceding turn, current turn, and session time. Query-intent extraction supplies
only the query. For both packet schedulers, the dynamic request uses the
candidate format
\texttt{[C\#\#\#] dia\_id=... speaker=... recorded\_date=...
content="..."} and explicitly requests the exact output cardinality. Answer
generation supplies the query and the materialized evidence blocks.

JSON responses are parsed directly and, when a provider wraps a requested list
in a generic object, the parser accepts the corresponding schema alias. Packet
scheduler outputs additionally undergo exact cardinality, uniqueness, and
closed-label validation. Judge responses are parsed into the two permitted
labels after substituting the question, gold answer, and generated answer into
Prompt~\ref{prm:judge}.

\begin{promptbox}{Fact Extraction System Message}
You extract high-value factual information from conversation turns. Return only a JSON array of complete standalone factual statements.
\end{promptbox}

%% file: custom.bib
@inproceedings{LLMASJUDGE,
 author = {Zheng, Lianmin and Chiang, Wei-Lin and Sheng, Ying and Zhuang, Siyuan and Wu, Zhanghao and Zhuang, Yonghao and Lin, Zi and Li, Zhuohan and Li, Dacheng and Xing, Eric and Zhang, Hao and Gonzalez, Joseph and Stoica, Ion},
 booktitle = {Advances in Neural Information Processing Systems},
 doi = {10.52202/075280-2020},
 editor = {A. Oh and T. Naumann and A. Globerson and K. Saenko and M. Hardt and S. Levine},
 pages = {46595--46623},
 publisher = {Curran Associates, Inc.},
 title = {Judging LLM-as-a-Judge with MT-Bench and Chatbot Arena},
 url = {https://proceedings.neurips.cc/paper_files/paper/2023/file/91f18a1287b398d378ef22505bf41832-Paper-Datasets_and_Benchmarks.pdf},
 volume = {36},
 year = {2023}
}

@inproceedings{papineni-etal-2002-bleu,
    title = "{B}leu: a Method for Automatic Evaluation of Machine Translation",
    author = "Papineni, Kishore  and
      Roukos, Salim  and
      Ward, Todd  and
      Zhu, Wei-Jing",
    editor = "Isabelle, Pierre  and
      Charniak, Eugene  and
      Lin, Dekang",
    booktitle = "Proceedings of the 40th Annual Meeting of the Association for Computational Linguistics",
    month = jul,
    year = "2002",
    address = "Philadelphia, Pennsylvania, USA",
    publisher = "Association for Computational Linguistics",
    url = "https://aclanthology.org/P02-1040/",
    doi = "10.3115/1073083.1073135",
    pages = "311--318"
}

@inproceedings{locomo,
    title = "Evaluating Very Long-Term Conversational Memory of {LLM} Agents",
    author = "Maharana, Adyasha  and
      Lee, Dong-Ho  and
      Tulyakov, Sergey  and
      Bansal, Mohit  and
      Barbieri, Francesco  and
      Fang, Yuwei",
    editor = "Ku, Lun-Wei  and
      Martins, Andre  and
      Srikumar, Vivek",
    booktitle = "Proceedings of the 62nd Annual Meeting of the Association for Computational Linguistics (Volume 1: Long Papers)",
    month = aug,
    year = "2024",
    address = "Bangkok, Thailand",
    publisher = "Association for Computational Linguistics",
    url = "https://aclanthology.org/2024.acl-long.747/",
    doi = "10.18653/v1/2024.acl-long.747",
    pages = "13851--13870"
}

@inproceedings{wu2024longmemeval,
  title        = {{LongMemEval}: Benchmarking Chat Assistants on Long-Term Interactive Memory},
  author       = {Wu, Di and Wang, Hongwei and Yu, Wenhao and Zhang, Yuwei and Chang, Kai-Wei and Yu, Dong},
  booktitle    = {The Thirteenth International Conference on Learning Representations},
  year         = {2025},
  url          = {https://openreview.net/forum?id=pZiyCaVuti},
  eprint       = {2410.10813},
  archivePrefix = {arXiv},
  primaryClass = {cs.CL},
  doi          = {10.48550/arXiv.2410.10813}
}

@misc{zep,
      title={Zep: A Temporal Knowledge Graph Architecture for Agent Memory}, 
      author={Preston Rasmussen and Pavlo Paliychuk and Travis Beauvais and Jack Ryan and Daniel Chalef},
      year={2025},
      eprint={2501.13956},
      archivePrefix={arXiv},
      primaryClass={cs.CL},
      url={https://arxiv.org/abs/2501.13956}, 
}

@inproceedings{Amem,
 author = {Xu, Wujiang and Liang, Zujie and Mei, Kai and Gao, Hang and Tan, Juntao and Zhang, Yongfeng},
 booktitle = {Advances in Neural Information Processing Systems},
 doi = {10.52202/085713-0593},
 editor = {D. Belgrave and C. Zhang and H. Lin and R. Pascanu and P. Koniusz and M. Ghassemi and N. Chen},
 pages = {17577--17604},
 publisher = {Curran Associates, Inc.},
 title = {A-Mem: Agentic Memory for LLM Agents},
 url = {https://proceedings.neurips.cc/paper_files/paper/2025/file/19909c36f51abc4856b4560aff3d36d6-Paper-Conference.pdf},
 volume = {38, Main Conference},
 year = {2025}
}

@misc{memorybank,
  title        = {{MemoryBank}: Enhancing Large Language Models with Long-Term Memory},
  author       = {Zhong, Wanjun and Guo, Lianghong and Gao, Qiqi and Ye, He and Wang, Yanlin},
  year         = {2023},
  eprint       = {2305.10250},
  archivePrefix = {arXiv},
  primaryClass = {cs.CL},
  doi          = {10.48550/arXiv.2305.10250}
}

@misc{mem0,
  title        = {Mem0: Building Production-Ready {AI} Agents with Scalable Long-Term Memory},
  author       = {Chhikara, Prateek and Khant, Dev and Aryan, Saket and Singh, Taranjeet and Yadav, Deshraj},
  year         = {2025},
  eprint       = {2504.19413},
  archivePrefix = {arXiv},
  primaryClass = {cs.CL},
  doi          = {10.48550/arXiv.2504.19413}
}

@misc{Atommem,
      title={AtomMem: Building Simple and Effective Memory System for LLM Agents via Atomic Facts}, 
      author={Yanyu Yao and Shangze Li and Zhi Zheng and Hui Zheng and Qi Liu and Tong Xu and Enhong Chen},
      year={2026},
      eprint={2606.19847},
      archivePrefix={arXiv},
      primaryClass={cs.CL},
      url={https://arxiv.org/abs/2606.19847}
}

@InProceedings{simplemem,
  title = 	 {{S}imple{M}em: Efficient Lifelong Memory for {LLM} Agents},
  author =       {Liu, Jiaqi and Su, Yaofeng and Xia, Peng and Han, Siwei and Zheng, Zeyu and Xie, Cihang and Ding, Mingyu and Yao, Huaxiu},
  booktitle = 	 {Proceedings of the 43rd International Conference on Machine Learning},
  pages = 	 {78500--78513},
  year = 	 {2026},
  editor = 	 {Zhang, Tong and Dudik, Miroslav and Jaggi, Martin and Agarwal, Alekh and Li, Sharon and Schuurmans, Dale and Zhu, Jerry and Berkenkamp, Felix and Dong, Hanze and Bietti, Alberto},
  volume = 	 {306},
  series = 	 {Proceedings of Machine Learning Research},
  month = 	 {06--11 Jul},
  publisher =    {PMLR},
  url = 	 {https://proceedings.mlr.press/v306/liu26dx.html}
}

@inproceedings{memoryos,
    title = "Memory {OS} of {AI} Agent",
    author = "Kang, Jiazheng  and
      Ji, Mingming  and
      Zhao, Zhe  and
      Bai, Ting",
    editor = "Christodoulopoulos, Christos  and
      Chakraborty, Tanmoy  and
      Rose, Carolyn  and
      Peng, Violet",
    booktitle = "Proceedings of the 2025 Conference on Empirical Methods in Natural Language Processing",
    month = nov,
    year = "2025",
    address = "Suzhou, China",
    publisher = "Association for Computational Linguistics",
    url = "https://aclanthology.org/2025.emnlp-main.1318/",
    doi = "10.18653/v1/2025.emnlp-main.1318",
    pages = "25961--25970",
    ISBN = "979-8-89176-332-6"
}

@inproceedings{anchormem,
    title = "{A}nchor{M}em: Anchored Facts with Associative Contexts for Building Memory in Large Language Models",
    author = "Shen, Zhanyu  and
      Cheng, Sijie  and
      Guo, Zhicheng  and
      Wang, Weiqin  and
      Wang, Yile  and
      Huang, Hui",
    editor = "Liakata, Maria  and
      Moreira, Viviane P.  and
      Zhang, Jiajun  and
      Jurgens, David",
    booktitle = "Findings of the {A}ssociation for {C}omputational {L}inguistics: {ACL} 2026",
    month = jul,
    year = "2026",
    address = "San Diego, California, United States",
    publisher = "Association for Computational Linguistics",
    url = "https://aclanthology.org/2026.findings-acl.1736/",
    doi = "10.18653/v1/2026.findings-acl.1736",
    pages = "34784--34798",
    ISBN = "979-8-89176-395-1",
}

@misc{xmemory,
      title={Beyond RAG for Agent Memory: Retrieval by Decoupling and Aggregation}, 
      author={Zhanghao Hu and Qinglin Zhu and Runcong Zhao and Di Liang and Hanqi Yan and Yulan He and Lin Gui},
      year={2026},
      eprint={2602.02007},
      archivePrefix={arXiv},
      primaryClass={cs.CL},
      url={https://arxiv.org/abs/2602.02007}, 
}

@misc{MemIR,
      title={Mitigating Provenance-Role Collapse in Long-Term Agents via Typed Memory Representation}, 
      author={Zhengda Jin and Bingbing Wang and Jing Li and Ruifeng Xu and Min Zhang},
      year={2026},
      eprint={2605.25869},
      archivePrefix={arXiv},
      primaryClass={cs.CL},
      url={https://arxiv.org/abs/2605.25869}, 
}

@inproceedings{fang2026lightmem,
  title        = {{LightMem}: Lightweight and Efficient Memory-Augmented Generation},
  author       = {Fang, Jizhan and Deng, Xinle and Xu, Haoming and Jiang, Ziyan and Tang, Yuqi and Xu, Ziwen and Deng, Shumin and Yao, Yunzhi and Wang, Mengru and Qiao, Shuofei and Chen, Huajun and Zhang, Ningyu},
  booktitle    = {The Fourteenth International Conference on Learning Representations},
  year         = {2026},
  url          = {https://openreview.net/forum?id=dyJ0GWpjJB}
}
